\documentclass[10pt,twocolumn,letterpaper]{article}

\usepackage{wacv_biblatex}              

\usepackage[T1]{fontenc} 
\usepackage{ifthen}
\usepackage{tabularx}
\usepackage{xr}
\newboolean{includeSupplemental}
\setboolean{includeSupplemental}{false}
\newboolean{useBiblatex}
\setboolean{useBiblatex}{true}
\usepackage{algorithm}
\usepackage{algorithmic}
\usepackage{multirow}

\definecolor{wacvblue}{rgb}{0.21,0.49,0.74}

\ifthenelse{\boolean{useBiblatex}}{
    \let\citep\undefined
    \let\citet\undefined

    \usepackage[style=ieee,backref=true,maxnames=3]{biblatex}
    \addbibresource{pi3_bibtex.bib}
    \usepackage[breaklinks,colorlinks,allcolors=wacvblue]{hyperref}

}

\begin{document}

\title{Improving MLLM Historical Record Extraction with Test-Time Image Augmentations}

\author{Taylor Archibald\\
Brigham Young University\\
Provo, UT, USA\\
{\tt\small tarch@byu.edu}
\and
Tony Martinez\\
Brigham Young University\\
Provo, UT, USA\\
{\tt\small martinez@cs.byu.edu}
}
\maketitle

\begin{abstract}
We present a novel ensemble framework that stabilizes LLM‑based text extraction from noisy historical documents. We transcribe multiple augmented variants of each image with Gemini 2.0 Flash and fuse these outputs with a custom Needleman–Wunsch–style aligner that yields both a consensus transcription and a confidence score. We present a new dataset of 622 Pennsylvania death records, and demonstrate our method improves transcription accuracy by 4 percentage points relative to a single‑shot baseline. We find that padding and blurring are the most useful for improving accuracy, while grid‑warp perturbations are best for separating high‑ and low‑confidence cases. The approach is simple, scalable, and immediately deployable to other document collections and transcription models.

\end{abstract}

\section{Introduction}
\label{sec:intro}
While traditional Optical Character Recognition (OCR) and Handwritten Text Recognition (HTR) systems have made significant strides, they often struggle with the variation of historical texts, such as severe degradation, diverse handwriting styles, and non-standard layouts~\cite{alkendiAdvancementsChallengesHandwritten2024}. The advent of Multimodal Large Language Models (MLLMs)—also known as Vision-Language Models (VLMs)—presents a new paradigm. Models such as GPT-4V \cite{GPT4VIsionSystem2024, yangDawnLMMsPreliminary2023} and Gemini~\cite{teamGeminiFamilyHighly2025} can perform challenging text extraction tasks in a zero-shot or few-shot manner, significantly reducing the need for extensive, task-specific training data~\cite{kimEarlyEvidenceHow2025, crosillaBenchmarkingLargeLanguage2025}.

However, despite early promising results, MLLMs are still susceptible to hallucination, struggle with noisy inputs, and provide no explicit or calibrated confidence score.  While a model's internal log probabilities are sometimes considered indicators of confidence, these scores can be poorly calibrated or inaccessible through proprietary APIs. Indeed, emerging research suggests that LLM-generated confidence scores often do not correlate well with accuracy and that models can exhibit overconfidence without a coherent internal sense of their own uncertainty \cite{pawitanConfidenceReasoningLarge2025}. This unreliability necessitates external validation mechanisms and motivates our work on an ensemble framework that derives confidence from agreement rather than relying on opaque internal model states.

\begin{figure}[t]
    \centering
    \begin{subfigure}[b]{0.23\textwidth}
        \includegraphics[width=\textwidth]{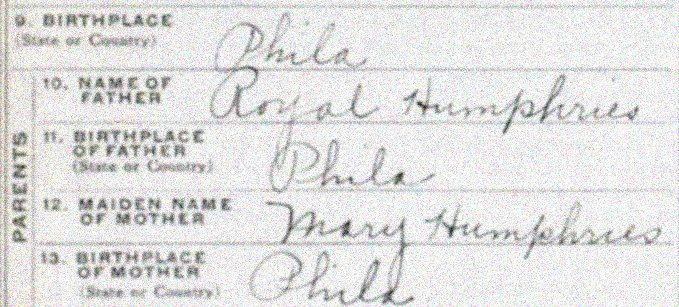}
        \caption{Noise}
        \label{fig:img_noise}
    \end{subfigure}
    \hfill 
    \begin{subfigure}[b]{0.23\textwidth}
        \includegraphics[width=\textwidth]{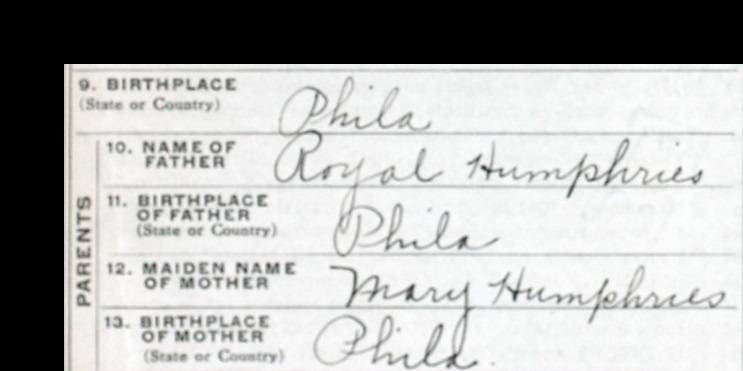}
        \caption{Padding}
        \label{fig:img_padding}
    \end{subfigure}


    \vspace{10pt} 

    \begin{subfigure}[b]{0.23\textwidth}
        \includegraphics[width=\textwidth]{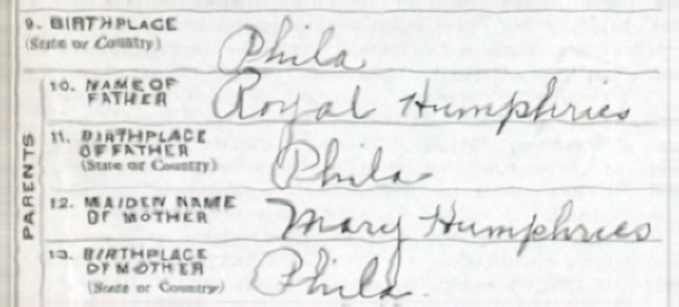}
        \caption{Grid Warp}
        \label{fig:img_gridwarp}
    \end{subfigure}
    \hfill 
    \begin{subfigure}[b]{0.23\textwidth}
        \includegraphics[width=\textwidth]{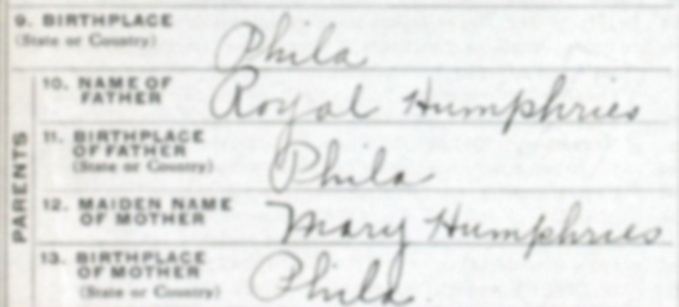}
        \caption{Blur}
        \label{fig:img_blur}
    \end{subfigure}

    \caption{A death record segment under different distortions can be ensembled to improve predictions and provide a confidence value.}
    \label{fig:2x2_layout_updated}
\end{figure}

To address this challenge, we introduce a scalable, black-box ensemble framework that uses a single MLLM to generate multiple transcriptions of a document image by applying a set of test-time augmentations (Fig.~\ref{fig:2x2_layout_updated}). By design, our method requires only one model, which avoids the complexities of managing distinct models that may have asymmetric costs, performance, or hardware requirements. We adapt the Needleman-Wunsch (NW) algorithm~\cite{needlemanGeneralMethodApplicable1970} to produce both a single consensus transcription and an agreement-based confidence score. To evaluate our method, we introduce a new dataset of 622 Pennsylvania death records, curated to ensure models have not previously trained on it. On this dataset, our method improves field transcription accuracy by over 4\% relative to a standard single-pass baseline. We assess different ensemble selection and augmentation strategies, and identify which transforms are most effective for improving accuracy or increasing confidence reliability.


\section{Related Work}
\label{sec:related}
Our approach builds upon recent advances in multimodal LLMs, traditional and modern ensemble methods, and confidence estimation.

\subsection{Multimodal LLMs for Document Understanding}

VLMs are increasingly applied to document analysis, combining visual understanding with language-based reasoning~\cite{xuLargeLanguageModels2024}. For example, LayoutLM learned joint representations of text and 2D layout information for scanned documents~\cite{xuLayoutLMPretrainingText2020} but still required external OCR model processing, while Donut and Dessurt achieved end-to-end document parsing without explicit OCR by directly predicting content from images via a transformer~\cite{kimOCRFreeDocumentUnderstanding2022,davisEndtoEndDocumentRecognition2023}.

More recently, the landscape has been further reshaped by the advent of large, more general-purpose VLMs or MLLMs, such as GPT-4V \cite{GPT4VIsionSystem2024, yangDawnLMMsPreliminary2023}, Gemini \cite{teamGeminiFamilyHighly2025}, and Claude 3\cite{IntroducingClaude3}. These demonstrate strong performance on OCR and information extraction tasks~\cite{kimEarlyEvidenceHow2025, crosillaBenchmarkingLargeLanguage2025}, particularly on contemporary printed or neatly handwritten text. However, their performance is diminished when faced with noisy, low-resource, or historical documents. In practice, complex layouts, dense script, degraded scans, or archaic scripts can lead even state-of-the-art models to misread characters or hallucinate content~\cite{biswasRobustnessStructuredData2024}.

\subsection{Confidence Score Generation}

Estimating the confidence of extracted text is crucial for downstream usability, especially with noisy historical data. While many traditional OCR systems provide confidence scores, LLMs do not have any built-in reliability checks, as models can produce plausible but incorrect outputs without indicating their uncertainty. Methods for estimating this uncertainty are often categorized as either white-box, requiring access to internal model states like output token probabilities, or black-box, which rely only on the model's textual output. Because many state-of-the-art LLMs are closed-source, white-box approaches are often not feasible, creating a need for effective black-box strategies. 


Among black-box strategies, a crucial insight revealed that prompting a model to verbally state its confidence can be a better-calibrated measure of accuracy than its internal probabilities~\cite{tianJustAskCalibration2023}. However, this approach of verbalized confidence can still be poorly calibrated and lead to overconfidence~\cite{xiongCanLLMsExpress2023}. Subsequent work like~\cite{borszukovszkiKnowWhatYou2025} confirmed overconfidence in VLMs when tested on corrupted image data, and found that as the severity of image corruption increased, verbalized confidence remained high despite worse performance.


\subsection{Ensemble Methods and Consensus Algorithms}

Combining outputs from multiple sources is a well-established technique to boost accuracy.
\begin{itemize}
    \item \textbf{Traditional Approaches:} These methods often rely on multiple, distinct OCR/HTR engines. ROVER uses dynamic programming to align outputs and vote to find the best hypothesis, often weighting votes by confidence~\cite{fiscusPostprocessingSystemYield1997}. Early success was demonstrated with this approach~\cite{loprestiUsingConsensusSequence1997}, and other work has combined HMM-based recognizers~\cite{bertolamiHiddenMarkovModelbased2008}. Ensemble voting has also been explored for post-OCR correction of character n-grams.
    \item \textbf{LLM-based Approaches:} LLMs enable new ensemble strategies. Self-consistency samples multiple reasoning paths from one model and takes a majority vote~\cite{wangSelfConsistencyImprovesChain2022}. Another approach involves combining outputs from different models~\cite{youngBenchmarkingMultipleLarge2025}. A method known as ConfBERT was recently proposed to integrate confidence scores from different OCR systems into a language model, significantly improving error spotting~\cite{hemmerConfidenceAwareDocumentOCR2024}. However, this approach still required each OCR model to produce a useful confidence score in the first place.
    
    Another black-box technique proposed generating a confidence score for OCR by measuring the agreement (entropy) between multiple *different* VLMs~\cite{zhangConsensusEntropyHarnessing2025}.
    However, relying on multiple distinct models introduces practical challenges related to the potential for disparate performance and operational costs. Moreover, it may be poorly adapted to shorter sequences of text (e.g., name fields), as our NW-consensus method consistently outperformed this alignment method on our dataset.

\end{itemize}

\subsection{Multi-Sample Consensus \& Test-Time Augmentation}
A key challenge for ensemble methods is generating multiple, diverse outputs without incurring the cost of running several different large models. One solution is test-time augmentation (TTA), a technique where a single trained model is applied to multiple augmented versions of an input instance during inference.

Creating multiple augmented views of an input at inference time and aggregating predictions is a known technique in computer vision~\cite{maTestTimeGenerativeAugmentation2024} and is being explored for LLMs~\cite{zhangSurveyTestTimeScaling2025}. TTA has been applied in various domains, including medical image segmentation, to improve model generalization and robustness to unseen variations~\cite{maTestTimeGenerativeAugmentation2024}. While this can improve robustness, its specific application to create ensembles for structured data extraction from historical documents using diverse image transforms on a single MLLM is not well established.

Our framework uniquely combines these threads: using diverse TTA with a single LLM for structured historical data extraction to generate both improved results and a calibrated confidence score, addressing key limitations of prior approaches.



\section{Method}
\label{sec:method}
Our approach enhances MLLM-based text extraction from historical documents by generating an ensemble of transcriptions from augmented input images. We apply mild distortions, such as padding, blurring, and grid-warping, to create multiple variants of each document scan. These variants are transcribed using an MLLM, and the resulting transcriptions are aligned using an adapted NW algorithm. By aggregating character-level votes across the ensemble, we produce a consensus transcription and a mathematically grounded confidence score.

\subsection{Formal Task Definition}

Let $\mathbf I\!\in\!\mathbb R^{H\times W\times C}$ be a document image drawn
from an unknown population $\mathcal P$, and let
$y\!\in\!\Sigma^{*}$ denote its latent ground‑truth transcription over an
alphabet~$\Sigma$.
Our goal is to learn an estimator $h$, that produces a transcription 
$\widehat y = h(\mathbf I)$ together with a scalar confidence
$\widehat p\!\in\![0,1]$ that minimizes the expected
\emph{character error rate} (CER)
\[
\mathcal L
=\mathbb E_{(\mathbf I,y)\sim\mathcal P}
  \bigl[\operatorname{CER}\bigl(h(\mathbf I),y\bigr)\bigr],
\qquad
\text{with }\widehat p\approx 1-\operatorname{CER}.
\]

\paragraph{Augmentation ensemble.}
Fix a finite grid of \emph{label‑preserving} augmentation settings
\(
\Theta_{\text{grid}}
  =\{\theta^{(1)},\dots,\theta^{(M)}\},
\)
spanning blur, noise, warp, and padding parameters
(Sec.~\ref{sec:augmentations}).
Label‑preserving means each operator
$\mathcal A_\theta$ satisfies
$\mathcal A_\theta(\mathbf I)\mapsto y$.
Draw $N$ distinct parameters
\(\theta_{1},\dots,\theta_{N}\) uniformly without replacement and form
\[
\mathbf I^{(n)} \;=\; \mathcal A_{\theta_{n}}(\mathbf I),
\qquad n=1,\dots,N.
\]

\paragraph{LLM extractor.}
A fixed multimodal MLLM
\(
\Gamma:\mathbb R^{H'\times W'\times C}\!\to\!\Sigma^{*}
\)
(e.g., Gemini) acts as a black‑box transcriber.  Applying
$\Gamma$ to each augmented image yields candidate strings
\(
\mathcal S(\mathbf I)=\{s_n\}_{n=1}^{N},
\;
s_{n}=\Gamma(\mathbf I^{(n)}).
\)

\paragraph{Consensus operator.}
A deterministic map
\(
\mathcal C:(\Sigma^{*})^{N}\!\to\!\bigl(\Sigma^{*},[0,1]\bigr),
\quad
\mathcal C(\mathcal S)=(\widehat y,\widehat p),
\)
implemented by our NW‑style voting aligner
(Sec.~\ref{sec:consensus}),
produces the final prediction
\(h=\mathcal C\!\circ\!\mathcal S\).

\subsection{Theoretical Perspective}

To gain theoretical insight into our ensemble method, we adopt a simplified statistical framework. We model the MLLM's error-generating process for a single character position~$k$ as a collection of correlated Bernoulli trials. Let $E_n \in \{0, 1\}$ be the error variable for the $n$-th transcription, where a majority vote fails if the sample mean of the errors, $\bar{E} = \frac{1}{N}\sum_{n=1}^N E_n$, exceeds $1/2$.

For this model to be tractable, we make two simplifying assumptions. First, we assume an average error probability, $\mathbb{E}[E_n] = \varepsilon_k < 1/2$, across the different augmentations. Second, we let~$\rho$ represent the average pairwise correlation between the error variables, acknowledging that the true dependency is more complex. Because the errors for $\{\mathbf{I}^{(n)}\}$ are not independent, this correlation inflates the variance of the sample mean:
\[
\text{Var}(\bar{E}) = \frac{\varepsilon_k(1-\varepsilon_k)}{N} \bigl[1 + (N-1)\rho\bigr].
\]
By comparing this to the variance for independent samples, we can define an \emph{effective sample size}, $N_{\text{eff}}$, which represents the number of independent samples that would yield an equivalent variance~\cite{kishSurveySampling1965}:
\[
N_{\text{eff}} = \frac{N}{1+(N-1)\rho}.
\]
The key insight from this model is that the ensemble behaves like a smaller sample of $N_{\text{eff}}$ independent voters. This allows us to adapt standard concentration inequalities to understand the probability of a majority vote error. For instance, using a Hoeffding-style analysis~\cite{hoeffdingProbabilityInequalitiesSums1994}, this framing suggests that the error probability has a bound that decays exponentially with the effective sample size:
\[
\Pr[\widehat y_k \neq y_k] \le \exp\bigl(-c \cdot N_{\text{eff}} \cdot \delta_k^2\bigr),
\qquad \text{where } \delta_k=\tfrac12-\varepsilon_k,
\]
and $c$ is a constant. This model provides the correct intuition: the guarantee correctly weakens as correlation increases ($\rho \to 1 \implies N_{\text{eff}} \to 1$). It therefore provides a formal motivation for our use of diverse, yet label-preserving, augmentations---their primary role is to minimize error correlation~$\rho$ and thus maximize the effective sample size of our ensemble.

\subsection{Dataset}

We utilize a newly contributed dataset comprising 622 scanned images of Pennsylvania death records from the early 20th century. This dataset presents significant challenges typical of historical archives, including varying handwriting styles, image degradation (stains, fading, noise), and inconsistent layouts. These documents have not been part of existing benchmarks, ensuring no risk of training data contamination for the MLLM.

The primary task is \textbf{structured information extraction}, specifically targeting the 3,684 non-blank name fields (\texttt{SelfGivenName}, \texttt{SelfSurname}, \texttt{MotherGivenName}, \texttt{MotherSurname}, \texttt{FatherGivenName}, and \texttt{FatherSurname}). We focus on these fields because they exhibit high variability and require no normalization (unlike dates or places). These fields are crucial for genealogical and historical research, making them a stringent test for extraction accuracy, though the same methods described here can generally be applied to most fields. 

\subsection{LLM and Ensemble Generation}

In our preliminary analysis, we found Gemini 2.0 Flash (Gemini) generally had the best performance among available MLLMs, which we use as the MLLM for our assessments. For each source document, we generate 20 distorted images for each distortion category and query the model once for each image, yielding 20 candidate extractions, each returned in a structured JSON format. For temperature experiments, we query the model 20 times on the unaltered image for the given sampling temperature. This approach operates entirely at test time, requires no model fine-tuning, and leverages the MLLM's sensitivity to input variations to explore the solution space.

\subsection{Data Augmentations}
\label{sec:augmentations}

To investigate the MLLM's response to various document image transformations, we employ 5 image-distortion strategies and 1 temperature variation strategy. For the image-distortion strategies, we conduct a grid search over 20 configurations and create an ensemble from the top 10. This selection is based on a 5-fold cross-validation performance on a dedicated validation set, with 10 samples generated, one for each chosen configuration.

While preliminary tests with Gemini showed good accuracy on the unaltered, high-resolution (median 3136×2904 pixels) images, we determined that processing at this native resolution was not essential for transcription and mostly incurred higher costs. We swept various resizing factors and found that a 50\% downscale was the smallest scale that performed no worse than the original. Consequently, we adopt 50\% resizing as a baseline post-processing step for most augmentations, as noted. The strategies we evaluate include:

\begin{enumerate}
\item \textbf{Blur + Resize}: This strategy combines Gaussian blurring using various odd-numbered pixel kernels (5--17px) with subsequent input to the MLLM at 100\%, 75\%, or 50\% of the original image resolution (omitting the 17px kernel for 50\% scale) to yield 20 distinct configurations.
\item \textbf{Resize Sweep}: This strategy directly explores the effect of image resolution. Images are resized to scale factors ranging from 30\% to 130\% of their original size, in 5\% increments, omitting the original size, to achieve 20 variations.
\item \textbf{Gaussian Noise}: This strategy simulates film grain or sensor noise by adding patch-based Gaussian noise to the full-resolution image, combining patch sizes of \{2, 4, 8, 16\} pixels with noise standard deviations (\(\sigma\)) of \{4, 6, 8, 10, 15\}, and then downscaling the image to 50\% of its original size.
\item \textbf{Pixel Shift Padding}: This strategy tests robustness to slight alignment changes by shifting the image content within a padded canvas using offset distances of \{8, 16, 32, 64, 128\} pixels in Northeast (\(\nearrow\)), Southeast (\(\searrow\)), Southwest (\(\swarrow\)), and Northwest (\(\nwarrow\)) directions, after which the resulting image is downscaled to 50\% of its original size.
\item \textbf{Grid Warp}: This strategy applies a smooth, non-linear distortion to the full-resolution image, simulating physical warping or creasing using mesh intervals of \{70, 85, 100, 115, 130\} pixels and distortion $\sigma$ of \{1, 2, 3, 4\}, based on principles similar to those in \cite{Wigington2018}, then downscaling the image to 50\% of its original size.
\item \textbf{Temperature}: This strategy investigates the MLLM's output diversity by querying it with the \textit{same} baseline image (downscaled to 50\% resolution), generating 20 samples each for temperatures of 0.5, 1.0, and 2.0, plus a single sample at temperature 0.0 as a quasi-deterministic reference, with each temperature setting constituting a separate experiment. For all temperature experiments, we set the cumulative probability threshold or top-p to 0.95. We posit that sampling with non-zero temperature can be used to approximate a white-box confidence estimation method, as both explore the model's output distribution.

\end{enumerate}

Notably, we found that even when temperature was set to 0, transcriptions were not completely deterministic, though the variation between runs was small (e.g., the CER differed by less than 0.1 percentage points).



\subsection{Finding Consensus}
\label{sec:consensus}
For each image, we have $k$ transcriptions obtained from distinct test‑time augmentations. Starting with the first string as a reference, we progressively align every new string to the current consensus with a NW~\cite{needlemanGeneralMethodApplicable1970} dynamic program ($match = +1$, $mismatch = -1$, $gap = -1$). Alignment columns carry a vote tally. After all $k$ strings are processed, the consensus character at column $j$ is the non‑gap symbol with the most votes; its confidence is that vote count divided by $k$. Word‑level confidence is the minimum of the character confidences in the word.


\begin{algorithm}[t]
\small
\caption{Progressive NW Consensus}
\label{alg:consensus}
\begin{algorithmic}[1]
\REQUIRE Transcriptions $\mathcal{S}=\{s_1,\dots,s_N\}$ from one image
\ENSURE Consensus string $\widehat{y}$ and per‑character confidences $\mathbf{p}$
\\[2pt]
\STATE $\widehat{y} \leftarrow s_1$ \COMMENT{initial consensus}
\STATE $\mathbf{V} \leftarrow$ list of dictionaries holding vote counts for each column of $\widehat{y}$
\FOR{$n=2$ \textbf{to} $N$}
    \STATE $(a_{\widehat{y}}, a_n) \leftarrow \text{NeedlemanWunsch}(\widehat{y},\; s_n)$
    \STATE $\mathbf{V} \leftarrow \textsc{UpdateVotes}(\mathbf{V}, a_{\widehat{y}}, a_n)$
    \STATE $\widehat{y} \leftarrow$ character‑wise $\arg\max$ votes in $\mathbf{V}$ (ignore gaps)
\ENDFOR
\STATE $\mathbf{p}[j] \leftarrow \mathbf{V}[j][\widehat{y}_j] / N \quad\forall j$
\RETURN $\widehat{y},\;\mathbf{p}$
\end{algorithmic}
\end{algorithm}

This approach extends classic methods like ROVER \cite{fiscusPostprocessingSystemYield1997} by incorporating dynamic gap/ambiguity symbols, explicit case handling, and importantly, an intrinsic confidence score that serves as a strong estimator of CER at inference time. Default hyperparameters were robust in cross-validation.

\subsection{Evaluation}

To assess the effectiveness of our ensemble framework, we evaluate both the accuracy of the transcriptions and the reliability of the confidence scores.

Our primary accuracy metric is the \textbf{Character Error Rate (CER)}, calculated as the Levenshtein distance between the predicted and ground truth transcriptions, normalized by the length of the ground truth. We also measure \textbf{Field Accuracy}, which is the percentage of transcriptions that are an exact match to the ground truth. For both metrics, we ignore case and exclude punctuation and spaces.

We evaluate performance at both the individual sample level and the consensus level. For a set of augmentations, we report the average CER and Field Accuracy across all individual transcriptions. We then report the same metrics for the single consensus transcription generated by our alignment algorithm.

To understand why certain ensembles are more effective than others, we measure the \textbf{Error Correlation} between pairs of transcriptions generated from different augmentations of the same source image. This is calculated as the Pearson correlation of the binary error vectors based on Field Accuracy (where 1 indicates an exact match failure and 0 indicates success) for each field across the dataset. Lower correlation suggests that augmentations are introducing diverse failure modes, which is desirable for an effective ensemble.

Finally, we assess the calibration of our agreement-based confidence scores. We use several standard metrics:
\begin{itemize}
    \item \textbf{Expected Calibration Error (ECE)} and \textbf{Adaptive Calibration Error (ACE)}, which measure the weighted average difference between confidence and accuracy across a number of bins.
    \item The \textbf{Brier Score}, which is the mean squared error between the predicted probability and the actual outcome.
    \item \textbf{Correlation} between confidence scores and the actual accuracy.
\end{itemize}
For these calibration metrics, we report the values for the raw confidence scores and also after applying \textbf{Isotonic Regression}, a non-parametric method to recalibrate the scores to better reflect the true likelihood of correctness.

\section{Results}
\label{sec:results}
\begin{table*}[!htbp]
\centering
\caption{Average and Consensus Character Error Rate, Field Accuracy, and Error Correlation by Transform Category}
\label{tab:enhanced_transform_metrics}
\begin{tabularx}{\textwidth}{lXXXXXX}
\toprule
\multirow{2}{*}{Transform} & \multicolumn{3}{c}{CER (\%)} & \multicolumn{2}{c}{Field Accuracy (\%)} & \multirow{2}{*}{Error Corr.} \\
\cmidrule(lr){2-4} \cmidrule(lr){5-6}
& Average & 5 Samples & 10 Samples & 5 Samples & 10 Samples & \\
\midrule
Baseline & 9.0 & 9.0 & 9.0 & 71.2 & 71.2 & --- \\
Temperature = 0.5 & 9.3 & 8.6 & 8.7 & 71.3 & 71.5 & 0.871 \\
Temperature = 1.0 & 10.2 & 9.1 & 8.8 & 69.9 & 70.4 & 0.780 \\
Temperature = 2.0 & 11.9 & 9.5 & 9.3 & 69.6 & 69.5 & 0.715 \\
Blur and Resize & 9.1 & 8.0 & 8.0 & 73.3 & 73.1 & 0.737 \\
Gaussian Noise & 9.3 & 8.1 & 8.0 & 72.7 & 73.3 & 0.745 \\
Grid Warp & 12.4 & 8.7 & 8.0 & 71.8 & 73.3 & \textbf{0.575} \\
Resize & \textbf{8.8} & 8.7 & 8.7 & 71.9 & 71.7 & 0.973 \\
Pad & 9.0 & \textbf{7.4} & \textbf{7.4} & \textbf{74.6} & \textbf{74.8} & 0.726 \\
\bottomrule
\end{tabularx}
\begin{flushleft}
\footnotesize
Baseline results are for 1-sample predictions (no consensus). Transform categories are sorted alphabetically. Lower error correlations indicate better diversity in failure modes.
\end{flushleft}
\end{table*}

\begin{figure}[h!]
    \centering
    \includegraphics[width=1.0\linewidth]{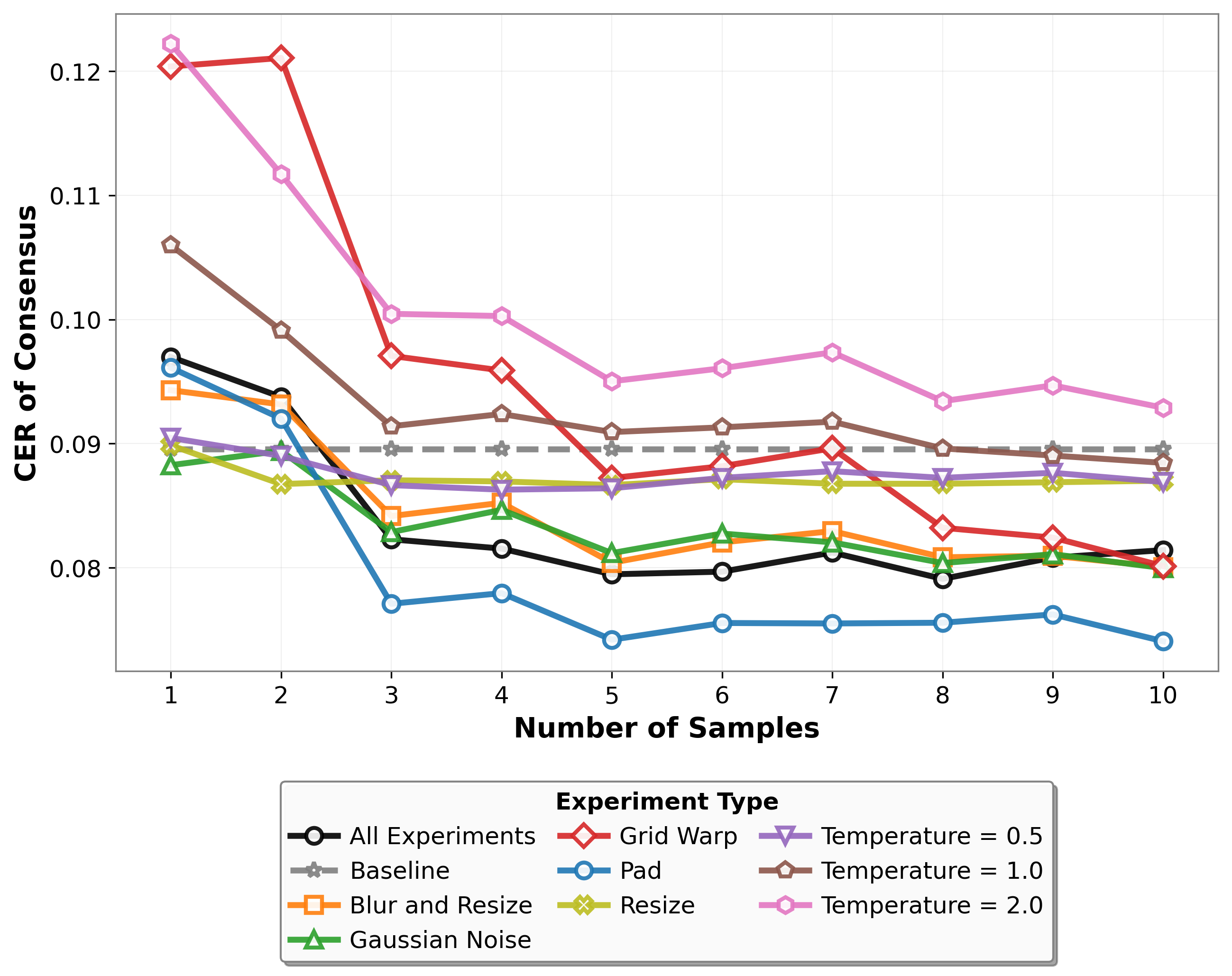}
    \caption{Consensus CER vs. Number of Samples by Experiment Type. This chart shows the performance of different augmentation strategies as more samples are added to the ensemble. While most methods improve on the baseline (dashed grey line), Pad, Blur and Resize, and Gaussian Noise consistently perform best.}
    \label{fig:experiment_type_chart}
\end{figure}

Our experiments demonstrate that the ensemble-based approach significantly improves transcription accuracy and provides meaningful confidence scores. We evaluated various augmentation strategies and ensemble selection methods to identify the most effective combinations.

\subsection*{Performance of Augmentation Strategies}

We first assessed the performance of different augmentation categories individually. Using 5-fold cross-validation, we selected the best-performing configurations from a validation set to form ensembles of 5 and 10. As shown in Table~\ref{tab:enhanced_transform_metrics}, nearly all augmentation strategies yield a consensus transcription that is more accurate than the single-pass baseline.

The \textbf{Pad} augmentation stands out, achieving the lowest consensus CER (7.4\%) and the highest Field Accuracy (74.8\% with 10 samples), an improvement of 1.6 percentage points in CER and 3.6 points in Field Accuracy over the baseline. \textbf{Blur and Resize} and \textbf{Gaussian Noise} also perform well. Interestingly, the \textbf{Grid Warp} augmentation, despite having the highest average CER (12.4\%) among individual samples, produces a highly effective consensus transcription (8.0\% CER), outperforming several other methods. 

This counterintuitive result is explained by its exceptionally low \textbf{Error Correlation} (0.575). The distortions introduced by Grid Warp cause the model to fail in diverse and uncorrelated ways, which makes the errors easier to correct during the consensus alignment. In contrast, simple \textbf{Resize} augmentations produce highly correlated errors (0.973), resulting in minimal improvement from ensembling. Figure~\ref{fig:experiment_type_chart} further illustrates that while most augmentations offer benefits over the baseline, the degree of improvement varies.

A notable portion of the ensemble's performance gain comes from successfully transcribing fields that the baseline model missed entirely. Across the dataset, the single-pass baseline failed to extract 7 name fields. Our ensemble methods, depending on the augmentation strategy, recovered between 4 and 7 of these. This improved field detection directly accounts for a 0.13 to 0.19 percentage point reduction in the overall Character Error Rate, showing the method enhances not just accuracy but also recall.

\subsection*{Ensemble Selection Strategy}

We also found that the method for selecting which augmentations to include in the ensemble is critical. Simply choosing the augmentations that perform best individually (based on validation CER) is suboptimal because their errors are often correlated.

\begin{figure}[h!]
    \centering
    \includegraphics[width=1.0\linewidth]{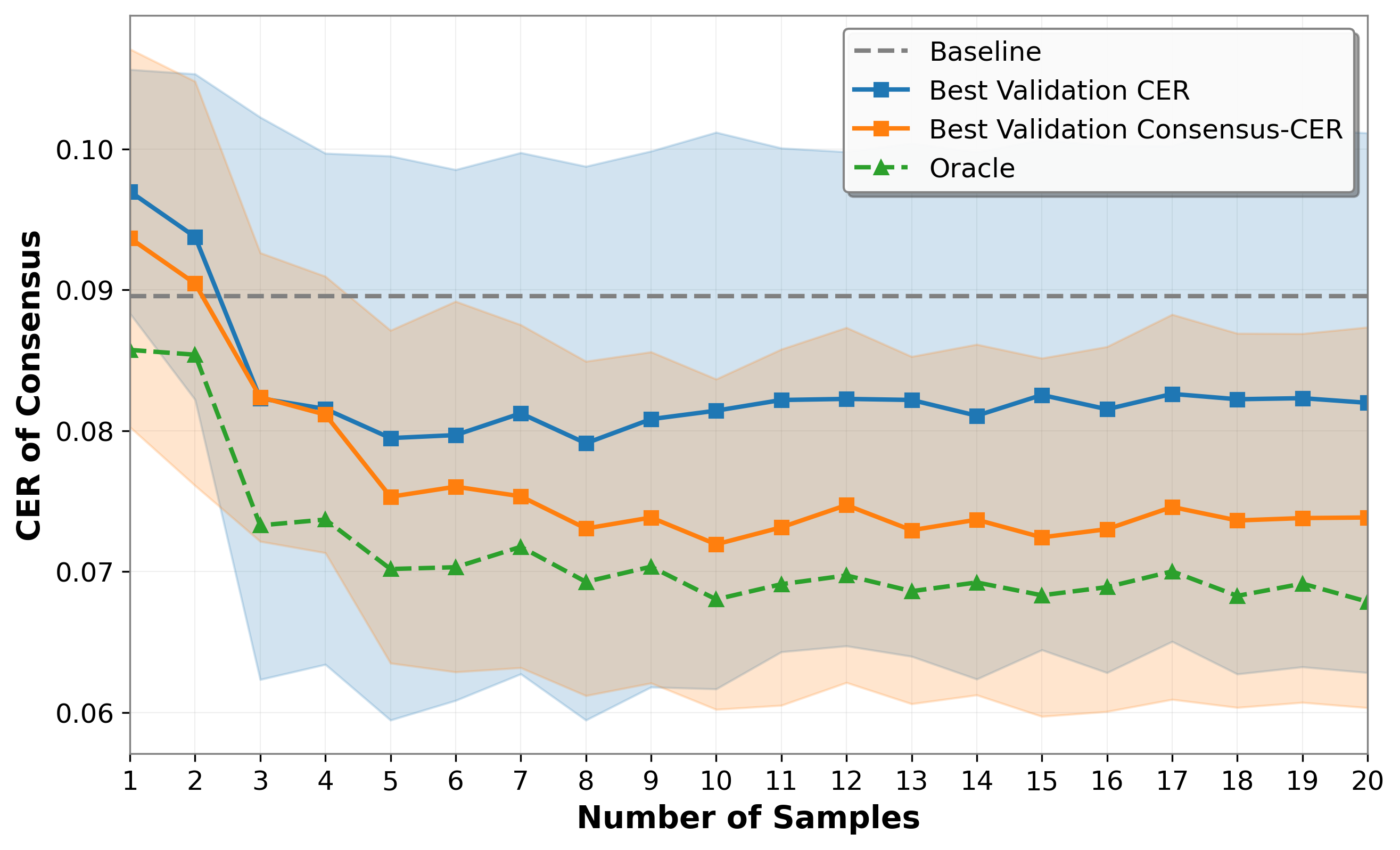}
    \caption{Comparison of Ensemble Selection Methods. This graph shows that selecting augmentations based on their contribution to the consensus accuracy (orange line) is superior to selecting them based on their individual accuracy (blue line) and approaches the theoretical best ``Oracle" performance (green dashed line).}
    \label{fig:ensemble_methods_comparison}
\end{figure}

Instead, a more effective approach is to select augmentations that most improve the \textit{consensus} transcription on the validation set. As detailed in Table~\ref{tab:ensemble_methods} and visualized in Figure~\ref{fig:ensemble_methods_comparison}, this ``Best Validation Consensus-CER" strategy outperforms the method of picking based on individual sample performance. With 10 samples, this method achieves a 7.2\% CER and a 75.2\% Field Accuracy, a \textbf{4 percentage point improvement} in accuracy over the 71.2\% baseline. This result approaches the performance of an ``Oracle" method that greedily selects the best possible augmentation at each step, indicating our selection strategy is highly effective.

\begin{table*}[!htbp]
\centering
\caption{Character Error Rate and Field Accuracy by Ensemble Method}
\label{tab:ensemble_methods}
\begin{tabularx}{\textwidth}{lXXXX}
\toprule
\multirow{2}{*}{Method} & \multicolumn{2}{c}{CER (\%)} & \multicolumn{2}{c}{Field Accuracy (\%)} \\
\cmidrule(lr){2-3} \cmidrule(lr){4-5}
& 5 Samples & 10 Samples & 5 Samples & 10 Samples \\
\midrule
Baseline & 9.0 & 9.0 & 71.2 & 71.2 \\
Oracle & 7.0 & 6.8 & 75.6 & 76.2 \\
Best Validation CER after Consensus & 7.5 & 7.2 & 74.6 & 75.2 \\
Best Validation CER & 7.9 & 8.1 & 73.5 & 73.3 \\
\bottomrule
\end{tabularx}
\end{table*}

\begin{table*}[!htbp]
\centering
\begin{tabular}{lcccccccccc}
\toprule
Transform & \# Exp & \# Rec & \multicolumn{2}{c}{Correlation} & \multicolumn{2}{c}{ECE} & \multicolumn{2}{c}{ACE} & \multicolumn{2}{c}{Brier} \\
\cmidrule(lr){4-11}
 & &  & Raw & Isotonic & Raw & Isotonic & Raw & Isotonic & Raw & Isotonic \\
\midrule
Blur and Resize & 20 & 3684 & 0.539 & 0.567 & 0.2188 & 0.0218 & 0.2182 & 0.0256 & 0.2138 & 0.1313 \\
Gaussian Noise & 20 & 3684 & 0.573 & 0.598 & 0.2225 & 0.0205 & 0.2225 & 0.0286 & 0.2166 & 0.1287 \\
Grid Warp & 20 & 3684 & \textbf{0.642} & \textbf{0.598} & \textbf{0.1677} & 0.0476 & \textbf{0.1677} & 0.0328 & \textbf{0.1746} & \textbf{0.1255} \\
Pad & 20 & 3684 & 0.521 & 0.569 & 0.2043 & 0.0167 & 0.2036 & 0.0319 & 0.2022 & 0.1302 \\
Resize & 20 & 3684 & 0.202 & 0.205 & 0.2758 & 0.0226 & 0.2755 & 0.0294 & 0.2748 & 0.1880 \\
Temperature = 0.5 & 12 & 3684 & 0.438 & 0.465 & 0.2663 & \textbf{0.0142} & 0.2663 & \textbf{0.0256} & 0.2600 & 0.1639 \\
Temperature = 1.0 & 12 & 3684 & 0.524 & 0.531 & 0.2510 & 0.0255 & 0.2505 & 0.0498 & 0.2437 & 0.1488 \\
Temperature = 2.0 & 12 & 3684 & 0.589 & 0.553 & 0.2402 & 0.0254 & 0.2402 & 0.0531 & 0.2316 & 0.1511 \\
\bottomrule
\end{tabular}
\caption{Reliability Analysis for Specific Transform Categories: Raw vs Isotonic Regression Calibration}
\label{tab:specific_categories_reliability}
\end{table*}

\subsection*{Confidence and Calibration}

A key advantage of our ensemble method is its ability to generate a calibrated confidence score for each transcription. We found that the \textbf{Grid Warp} augmentation, which creates diverse errors, is particularly effective for calibration. As shown in Table~\ref{tab:specific_categories_reliability}, Grid Warp achieves the highest raw correlation (0.642) between confidence and accuracy and is a strong choice for generating reliable confidence estimates directly.

\begin{figure}[h!]
    \centering
    \includegraphics[width=1.0\linewidth]{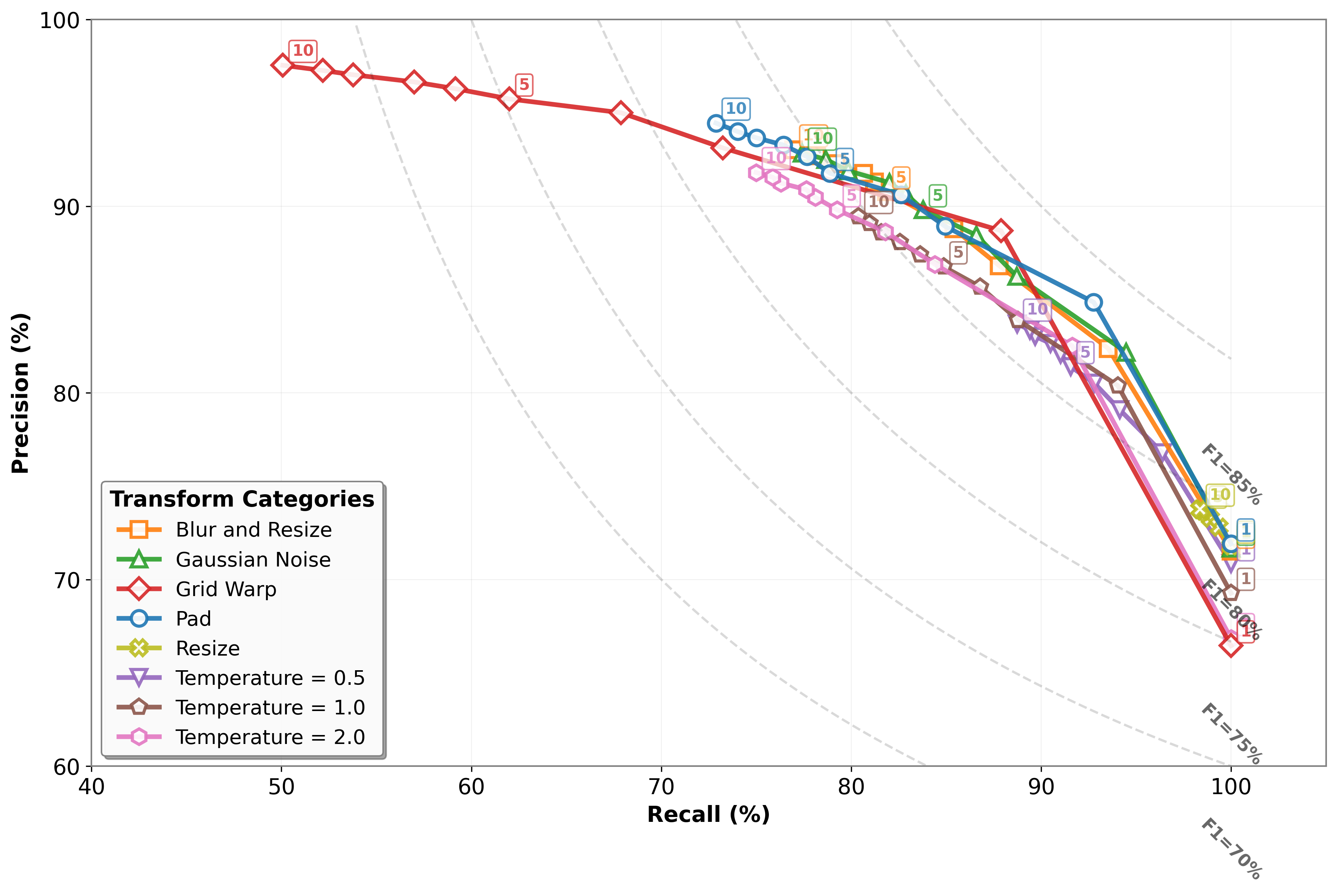}
    \caption{Precision-Recall for Unanimous Predictions. This chart shows the tradeoff between precision and recall for different augmentation categories. Pad (blue) achieves a high F1-score, while Grid Warp (red) offers higher precision, especially with more samples (indicated by the number labels).}
    \label{fig:precision_recall_chart}
\end{figure}

The precision-recall tradeoff, visualized in Figure~\ref{fig:precision_recall_chart}, further highlights the strengths of different augmentations. Here, we define a ``positive" as a field where the ensemble reached a unanimous prediction. The \textbf{Pad} augmentation achieves the highest F1-score, indicating a good balance between precision and recall. However, \textbf{Grid Warp} demonstrates superior precision, particularly as more samples are added to the ensemble. This means that when the ensemble based on \textbf{Grid Warp} augmentations reaches a unanimous agreement, that prediction is highly likely to be correct. This is valuable for workflows where high-confidence transcriptions can be automatically accepted, while low-confidence ones are flagged for human review.

\subsection*{Qualitative Analysis of Errors}

\begin{figure}[h!]
    \centering
    \begin{subfigure}[b]{0.45\textwidth}
        \centering
        \includegraphics[width=\textwidth]{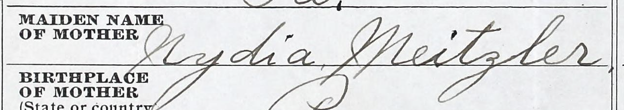}
        \caption{\textbf{Ambiguous characters}: The model consistently predicts ``Lydia," though the ground truth is ``Nydia".}
        \label{fig:nydia_error}
    \end{subfigure}
    \hfill 
    \begin{subfigure}[b]{0.45\textwidth}
        \centering
        \includegraphics[width=\textwidth]{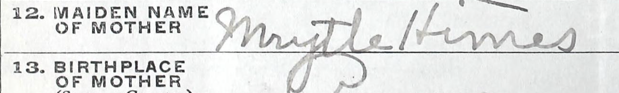}
        \caption{\textbf{Spelling autocorrect}: The model ``corrects" the spelling of ``Mrytle" to ``Myrtle."}
        \label{fig:myrtle_error}
    \end{subfigure}
    \hfill 
    \begin{subfigure}[b]{0.45\textwidth}
        \centering
        \includegraphics[width=\textwidth]{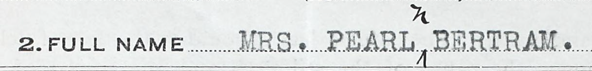}
        \caption{\textbf{Additions and Deletions}: The middle initial ``N" is consistently missed.}
        \label{fig:pearl_error}
    \end{subfigure}
    \caption{Across 3,684 predicted name fields, 870 had consistent predictions using the full pool of 100 augmentation configurations. We present 3 of the 7 times where the ensemble was unanimous but inconsistent with our ground truth, based on an extended analysis using the full pool of 100 augmentation configurations.}
    \label{fig:model_weaknesses}
\end{figure}

Despite the significant accuracy gains, our ensemble method is not infallible. We analyzed the rare cases where the ensemble was unanimously confident but still incorrect. These errors typically fall into three categories, as illustrated in Figure~\ref{fig:model_weaknesses}:
\begin{enumerate}
    \item \textbf{Ambiguous Characters:} Certain handwritten characters are difficult to distinguish, and the model may consistently favor the more common word.
    \item \textbf{Spelling Autocorrect:} The model sometimes ``corrects" unusual but accurate spellings (e.g., ``Mrytle" to ``Myrtle"), sacrificing literal accuracy for what it perceives as correctness.
    \item \textbf{Additions and Deletions:} Struck-through text or small, interlineated additions are often missed by the model, leading to omissions in the transcription.
\end{enumerate}
These failure modes highlight the remaining challenges in transcribing complex historical documents and suggest areas for future improvement in MLLM training and fine-tuning.

\section{Conclusion and Future Work}
\label{sec:discussion}
We have presented a simple yet effective ensemble framework that improves the reliability of LLM-based text extraction from noisy historical documents. By generating and aligning transcriptions from multiple augmented images, our method boosts accuracy by 4 percentage points and, crucially, produces a well-calibrated confidence score without accessing internal model states. Our analysis revealed that specific augmentations serve distinct purposes: padding is most effective for improving accuracy, while grid-warp's ability to introduce diverse errors is ideal for robust confidence estimation. The approach is black-box, scalable, and readily applicable to other document collections.

Several avenues for future work remain. The consensus mechanism could be refined beyond simple majority voting to a weighted scheme, perhaps giving more influence to predictions from less severe augmentations or by combining different degradation types into a single, optimized ensemble. The framework's effectiveness should also be validated on a broader range of documents, such as long-form letters or journal collections, which present different challenges in layout and content.

A particularly promising research direction involves a deeper analysis of why pixel-shift padding is so effective. A border of 8–128 pixels shifts the entire page, forcing Gemini 2.0 Flash to re-tile the image into blocks before upsampling. Each tile is then further patchified, typically into 16×16 tokens. Large pixel shifts therefore expose new content at tile boundaries and alter dozens of patch offsets, creating “token jitter” that diversifies the model’s predictions without harming readability. This aligns with evidence that smaller patches can retain more information \cite{wangScalingLawsPatchification2025, touvronThreeThingsEveryone2022}. Future work should (i) log the exact tile grid under controlled shifts to better understand this mechanism, (ii) sweep patch sizes in a native-resolution Vision Transformer to measure information loss \cite{qiaoUniViTARUnifiedVision2025}, and (iii) attempt to learn a content-aware padding policy that jointly optimizes both tiling and patchification.

\section*{Acknowledgment}

The authors would like to thank the Handwriting Recognition Team at Ancestry.com for providing the dataset for this study.


\ifthenelse{\boolean{useBiblatex}}{
    \small
    \clearpage
    \printbibliography
    \newpage
}{
}


\end{document}